\documentclass{article}

\usepackage{arxiv}

\usepackage{graphicx}
\usepackage{subcaption}
\usepackage[utf8]{inputenc} 
\usepackage[T1]{fontenc}    
\usepackage{hyperref}       
\usepackage{url}            
\usepackage{booktabs}       
\usepackage{amsfonts}       
\usepackage{nicefrac}       
\usepackage{microtype}      
\usepackage{cite}

\title{Precipitation nowcasting using a stochastic variational frame predictor with learned prior distribution}

\author{
  Alex Bihlo \\
  Department of Mathematics and Statistics\\
  Memorial University of Newfoundland\\
  \texttt{abihlo@mun.ca}
}

\begin{document}
\maketitle

\begin{abstract}
We propose the use of a stochastic variational frame prediction deep neural network with a learned prior distribution trained on two-dimensional rain radar reflectivity maps for precipitation nowcasting with lead times of up to 2 1/2 hours. We present a comparison to a standard convolutional LSTM network and assess the evolution of the structural similarity index for both methods. Case studies are presented that illustrate that the novel methodology can yield meaningful forecasts without excessive blur for the time horizons of interest. 
\end{abstract}

\keywords{Precipitation nowcasting \and deep learning \and variational autoencoder \and generative modeling}

\section{Introduction}

Accurate short-range (0 to 6 hours) precipitation forecasts are of great practical importance for various stakeholders, including airport authorities, event planers and insurance companies. While long-range precipitation forecasts are traditionally obtained from numerical weather prediction models, short-term forecasts are more challenging to obtain. The shorter lead times and higher resolutions required do not mesh well with traditional numerical weather predictions, which are costly to obtain and might not capture the initial precipitation distribution correctly, despite much recent improvements owing to increased computational power and improved forecasting strategies. 

On the very short time scale typically other strategies are being used, with the most prominent one being extrapolation based methods. These methods generally compute an optical flow based on the last few precipitation fields by calculating an approximate flow velocity and then using semi-Lagrangian advection to move areas of precipitation along the calculated flow field, see e.g.~\cite{bowl04a}. A main complication of these methods is that they require appropriate pre-processing of the raw radar image data, such as smoothing of the input data or image segmentation to label contiguous areas of precipitation. 
As an alternative to these methods, recent years have seen an increased interest in machine learning based strategies. These methods have the advantage of being end-to-end trainable, with minimal, if any, pre-processing required. 

Here we propose a novel deep learning approach to precipitation nowcasting. We implement a variation of the model proposed by~\cite{dent18a}, which was used in the field of video frame prediction. The method employs an encoder--prediction--decoder framework which resembles a conditional variational autoencoder~\cite{king13a}, whose latent space prior distribution is learned based upon previous video frames. We then train the proposed model on a short series of radar frames with the aim to forecast the precipitation over the next several hours.

\section{Related work}\label{sec:RelatedWork}

The precipitation nowcasting problem has seen considerable interest in the meteorological community, with several decades of research spent on formulating increasingly refined extrapolation based methods. A main reason for the prevalence of extrapolation based methods compared to direct numerical weather predictions is that numerical models generally do not capture the initial precipitation well~\cite{lin05a}. Various different methods using extrapolation-based approaches can be found in~\cite{bowl04a,lin05a,zahr12a}, as well as in the references therein.

Here we are exclusively concerned with machine learning methodologies, in particular those based on deep learning. This is a relatively recent application field of machine learning and a fundamentally different approach to the precipitation nowcasting problem compared to the so far dominant numerical or extrapolation based approaches. Recent examples for the deep learning methodology using convolutional recurrent neural networks can be found in the seminal paper~\cite{xing15a}, as well as in~\cite{shi17a}.

It is important to stress that precipitation forecasting based on radar images is an ill-posed problem: \emph{The radar reflectivity maps do not contain the entire physically relevant information to uniquely forecast the rainfall rates in future radar maps.} In this sense, the precipitation forecasting problem can be regarded as a particular case of video frame prediction, which is also an ill-posed problem. There are multiple possible solutions to this problem and a machine learning system aims to find the most probable ones.

Several approaches to video frame prediction using deep learning have been proposed in recent years, in particular methods based on LSTM autoencoders-like networks~\cite{baba17a,dent18a,finn16a,sriv15a}, generative adversarial network~\cite{lee18a,math15a,vond17a} and probabilistic approaches~\cite{xue16a}.

Here we adopt a slight modification of the methodology proposed in~\cite{dent18a}, see also~\cite{baba17a,lee18a} for similar approaches. In particular, we train a stochastic variational autoencoder-like architecture with a learned prior distribution.

\section{Methodology}

The description of the model used here follows~\cite{dent18a}. We use a prediction model $p_{\rm pred}$ within an encoder--decoder framework that maps the single given input radar image $\mathbf{x}_{i-1}$ from physical space to latent space, where it is passed through a stack of convolutional LSTM layers, and back to physical space, yielding $\hat\mathbf{x}_i$ aiming to closely approximate the next true radar image $\mathbf{x}_i$. 

The prediction in latent space is conditioned upon a random vector $\mathbf{z}_t$ that is sampled from a multivariate Gaussian prior distribution $p_{\rm prior}(\mathbf{z}_t | \mathbf{x}_{1:i-1})$, that is learned from the previous series of $i-1$ input frames, $\mathbf{x}_{1:i-1}$. The main idea is that the learned prior distribution encodes important stochastic information about the next frame that is not included in the deterministic prediction model. For the precipitation nowcasting problem, we expect the prior model to learn the basic physical rules by which precipitation cells evolve and move in time and space. 

The prior model distribution $p_{\rm prior}$ is trained with a separate inference model, both in practice being recurrent neural networks, that takes as input the target frame $\mathbf{x}_{i}$ and aims to compute the distribution $q_{\rm inf}(\mathbf{z}_i | \mathbf{x}_{1:i})$, where the dependence on the previous frames $\mathbf{x}_{1:i-1}$ is due to the recurrent nature of the neural network structure for the inference model chosen. 

The similarity between the distribution $q_{\rm inf}(\mathbf{z}_i | \mathbf{x}_{1:i})$ and the prior distribution $p_{\rm prior}(\mathbf{z}_i | \mathbf{x}_{1:i-1})$ is enforced by including a Kullbeck--Leibler divergence term in the cost function to be optimized. Note that the prior distribution depends on the previously seen radar images, but does not include the frame which will be predicted next. 

The model optimizes the variational lower bound
\[
 \mathcal L(\mathbf{x}_{1:T}) = \sum_{i=1}^T \left(\mathbb{E}_{q_{\rm inf}(\mathbf{z}_{1:i}|\mathbf{x}_{1:i})} \log p_{\rm pred}(\mathbf{x}_i|\mathbf{x}_{1:i-1},\mathbf{z}_{1:i})-\beta \mathrm{D}_{\rm KL}(q_{\rm inf}(\mathbf{z}_i|\mathbf{x}_{1:i})|| p_{\rm prior}(\mathbf{z}_i|\mathbf{x}_{1:i-1}))\right),
\]
which will jointly train the encoder--prediction--decoder network as well as the prior model network. The first term in the cost function corresponds to the reconstruction loss (in our case simply the $l_2$-loss between predicted and observed radar images), the second term is the Kullback--Leibler divergence. Note that this cost function is different from the standard variational autoencoder for which $\beta=1$. As was found in \cite{higg17a}, the parameter $\beta$ allows for tuning the balance between the latent space capacity and the reconstruction accuracy.

\begin{figure}[!ht]
    \centering
    \begin{subfigure}[b]{0.49\textwidth}
        \includegraphics[scale=0.45]{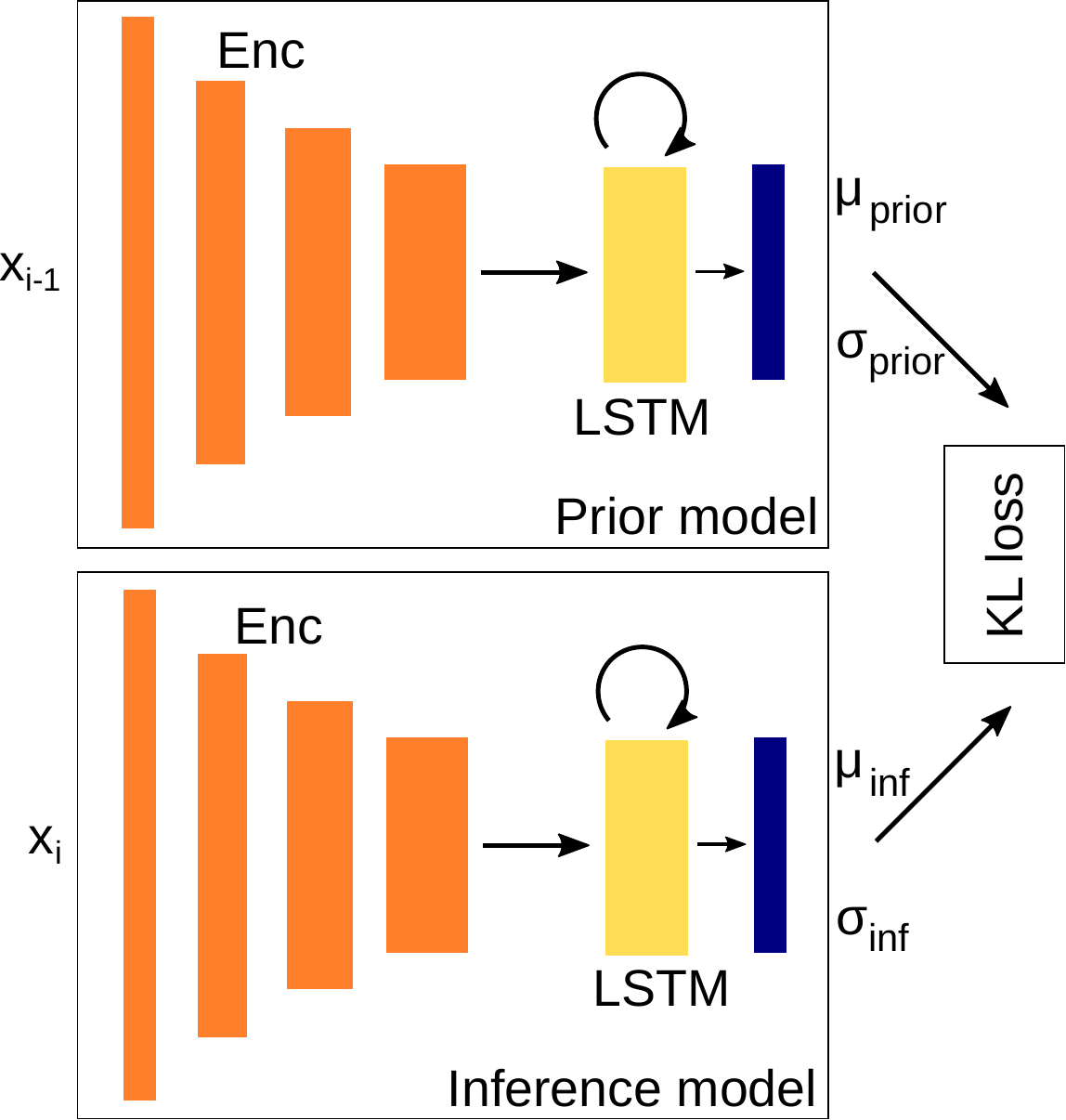}
        \caption{Prior model training step.}
        \label{fig:prior}
    \end{subfigure}
    \hspace{-1.5cm}
    \begin{subfigure}[b]{0.49\textwidth}
        \includegraphics[scale=0.45]{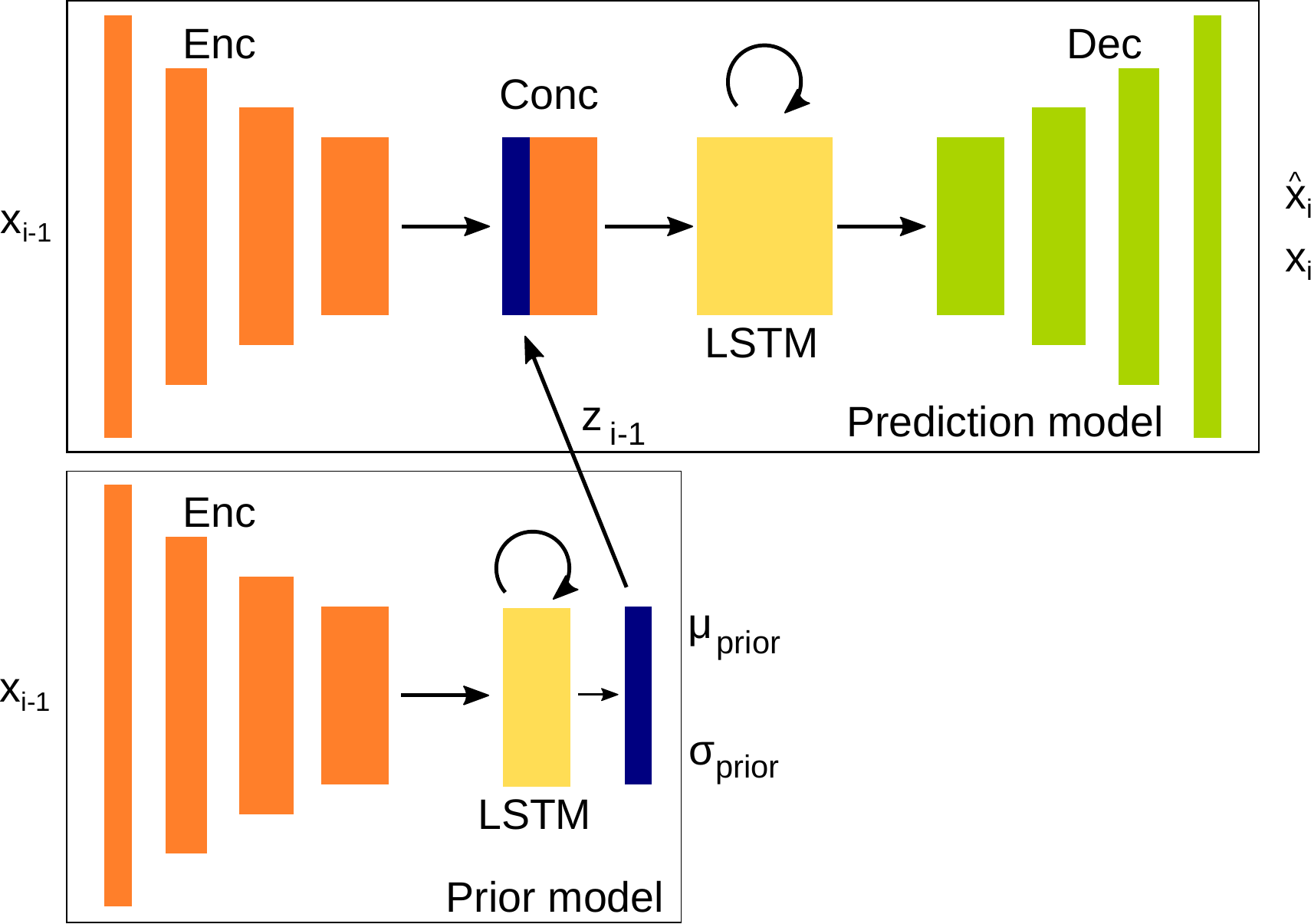}
        \caption{Prediction model training/test step.}
        \label{fig:predictionStep}
    \end{subfigure}
    \caption{Training of the stochastic variational frame predictor framework. \emph{Left:} Training of the prior model is done by minimizing the KL divergence term. \emph{Right:} At \emph{training} time the optimizer attempts to minimize the $l_2$ reconstruction error based on the true frames $\mathbf{x}_i$ and the predicted frames $\hat\mathbf{x}_i$, balanced with minimizing the KL divergence. At \emph{test} time the prior network learns stochastic information from the series of previous images $\mathbf{x}_{1:i-1}$ and passes this information, along with $\mathbf{x}_{i-1}$, into the deterministic prediction network to predict $\hat\mathbf{x}_i$.}\label{fig:predictionModel}
\end{figure}

The general model architecture is represented in Figure~\ref{fig:predictionModel}.

The difference between our model and the one developed in~\cite{dent18a} is that we use convolutional LSTM layers for the prediction model only and regular LSTM layers for the inference and the prior model rather than convolutional LSTM layers for all model components. We also do not implement skip connections between the last encoder and first decoder layer since our backgrounds typically correspond to areas of zero precipitation which is readily learned by the model without using skip connections.

\section{Model}

\subsection{Architecture}

The encoder network consists of four stacked, stride-2 convolutional layers with 16, 32, 64, and 128 filters, respectively. The filter size is $5\times 5$ for the first two layers and $3\times 3$ for the last two layers. No pooling or dense layers are used in the encoder and decoder networks. The latent space prediction model consists of two stacked convolutional LSTM layers with $128$ filters and a filter size of $3\times 3$ each. The inference and prior model add an extra stride-2 convolutional layer with 128 filters and filter size of $3\times 3$ before a single conventional LSTM with 64 neurons and a densely connected layer with linear activation and 70 neurons, which corresponds to the dimension of our latent space.

The model is trained on $n_{\rm i}=5=T-1$ radar images each, sampled every $15$ minutes, with the aim to forecast the next $n_{\rm p}=10$ images, i.e.\ up to 2 1/2 hours in the future. Each radar frame is forecast separately and the multi-step forecast is accomplished by including the generated frames back into the input stack of the prediction model.

The model is implemented in \texttt{Python3} using \texttt{Keras}~\cite{chol15a} with \texttt{TensorFlow} backend. Initial training was done using one NVIDIA T4 GPU on the interactive \texttt{Google Colab} platform, production runs were carried out using one NVIDIA P100 GPU on the Compute Canada cluster \texttt{Graham}. 

Training was done using the Adam optimizer~\cite{king14a} with a default learning rate of $\eta = 0.001$. The parameter $\beta$ was set to $\beta=10^{-7}$, which corresponded to the optimal value tested in our hyper-parameter studies within the range of $\beta\in[10^{-8},1]$. This value is consistent with the one used in~\cite{dent18a}. The training data size consisted of approximately 10,000 samples of six frames each (5 input frames, 1 prediction frame). A total of 10\% of the training data was held out for validation. The test data consisted of 1,000 samples (5 input frames, 1 prediction frames; multi-frame predictions were obtained by feeding the predicted image back into the stack of 5 input frames). Early stopping was employed in all experiments.

\subsection{Data}

The proprietary radar data was provided by Austro Control, the Austrian air navigation service provider, and consists of two-dimensional reflectivity maps recorded by four weather radar stations (Feldkirchen, Patscherkofel, Rauchenwarth, Zirbitzkogel) over the Austrian region. The original resolution of the data is 1 km $\times$ 1 km recorded every $5$ minutes, but has been down-sampled to 5 km $\times$ 5 km (giving image sizes of $160\times 110$ pixels) every $15$ minutes to allow for reasonably fast training times, and for mini-batches of images to still fit into the memory of a single GPU. The data consists of converted rain rates, which employs the standard $Z$--$R$ (radar reflectivity to rain rate) relation of the form $Z=AR^b$ with the empirical constants $A=200$ and $b=1.6$ according to Marshall and Palmer~\cite{marsh48a}, see also~\cite{kalt15a,wils79a}. The converted rain rates (in mm/h) fall into $14$ classes, which our method aims to predict in an end-to-end fashion. For more information on the radar data, see~\cite{kalt15a}.

A total of 5 years of data, from January 2014 to December 2018 has been used in this studies, with 4 years being used for training and 1 year for testing. To remove cases with little to no precipitation we require that the cumulative rain rate over the entire image domain is at least 10,000 mm/h for each valid training and test case.

\section{Results}

As a baseline comparison we use a two-layer convolutional LSTM model with $64$ filters of size $3\times 3$, similar to the one that was found in~\cite{xing15a} to give the best results for the precipitation nowcasting problem. We have tested other two-layer (using $72$ filters each) and three-layer (using $64$ filters each) convolutional LSTM architectures as well, which generally yielded worse results than those presented, possibly due to over-fitting. 

Note that the convolutionsl LSTM models could be trained to predict the next $n_{\rm p}$ frames in one sweep based on the given $n_{\rm i}$ frames. We abstain from doing this and rather predict one frame at the time and feed the predicted frame back in the stack of input frames. The reason for doing this is that the prediction in one sweep would not allow for a coupled dynamics among subsequent image frames to develop over the course of the prediction horizon, in contrast to the case when each individually predicted frame is fed back into the stack of input frames. The latter possibility is also more flexible in that we can choose the prediction horizon to be as long as we wish, which is an appealing practical feature for precipitation nowcasting with various lead times. 

\begin{figure}[!ht]
    \centering
    \includegraphics[scale=0.5]{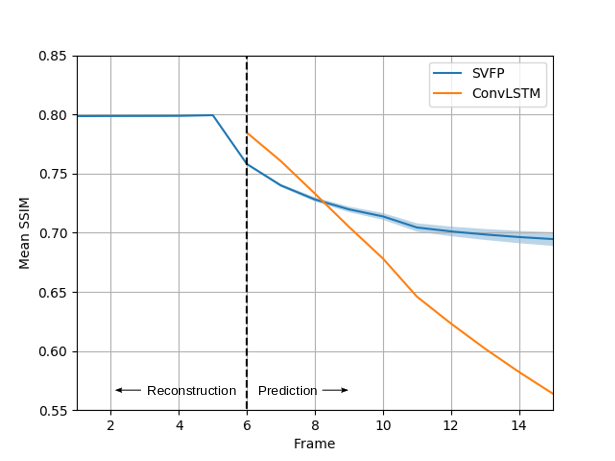}
    \caption{Mean SSIM for the stochastic variational frame predictor network with 2 layer, 128 filters each prediction network (95\% confidence interval is shaded) and standard convolutional LSTM network with 2 layer, 64 filters each. Prediction starts at frame 6. Frames 1 to 5 for the SVFP model show the ability of input reconstruction.}
    \label{fig:SSIM}
\end{figure}

We compute the structural similarity index (SSIM) over the test data for each of the 10 frames for every prediction. The time series of the mean over the test data set is presented in Figure~\ref{fig:SSIM}. For the sake of completeness, we also show the reconstruction quality of the conditional variational autoencoder method (frames 1 to 5). The prediction starts at frame 6. Figure~\ref{fig:SSIM} illustrates that the vanilla convolutional LSTM network gives better predictions for the first two frames but its forecast quality drops significantly afterwards. The stochatistic variational method in turn gives better longer term forecasts, with the drop in SSIM being significantly less compared to the convolutional LSTM network. This is consistent with the observation that convolutional LSTM networks tend to blur later frames when used in frame-by-frame video prediction, which was one main motivation for the stochastic variational frame prediction method being introduced in~\cite{dent18a}. In the present case, this translates to sharper predictions of precipitation cells with longer lead times, an attractive feature for the real-world nowcasting problem.


\begin{figure}[!ht]
    \centering
    \begin{subfigure}[b]{\textwidth}
        \centering
        \includegraphics[scale=1.3]{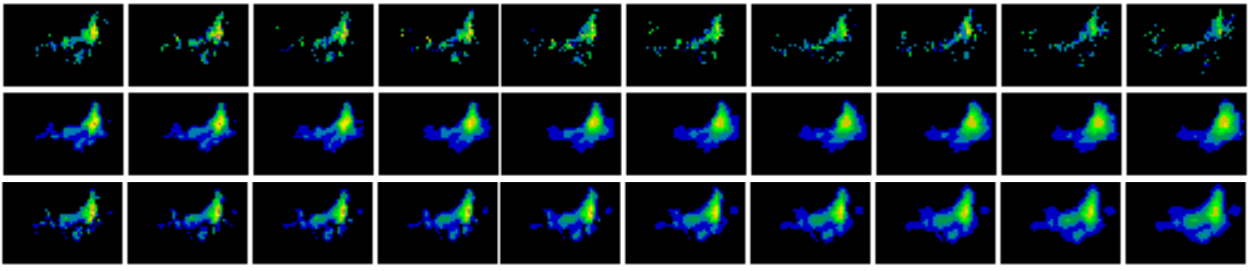}%
    \end{subfigure}
    ~
    \begin{subfigure}[b]{\textwidth}
        \includegraphics[scale=1.3]{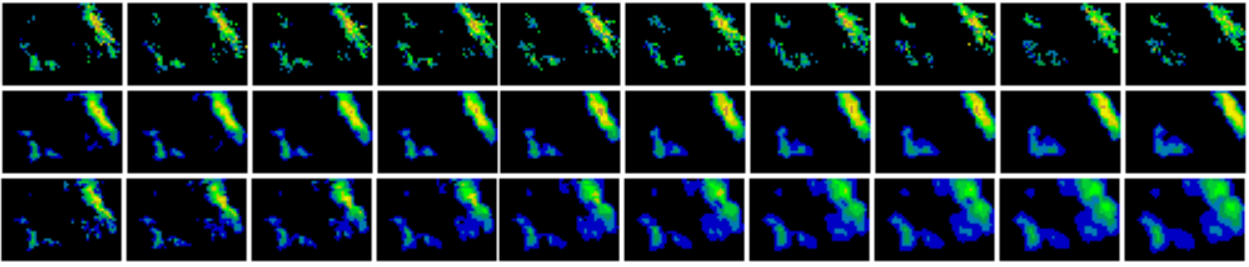}%
    \end{subfigure}
    ~
    \begin{subfigure}[b]{\textwidth}
        \includegraphics[scale=1.3]{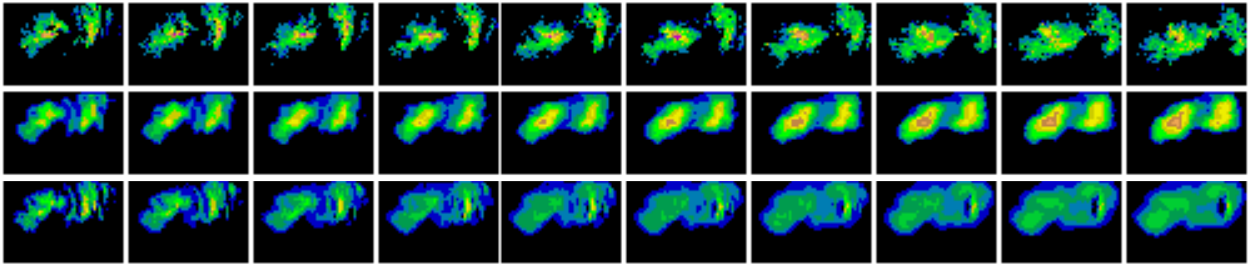}%
    \end{subfigure}
        ~
    \begin{subfigure}[b]{\textwidth}
        \includegraphics[scale=1.3]{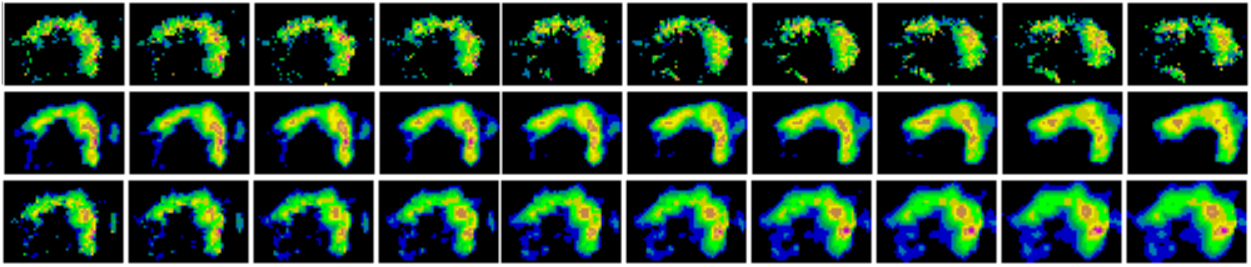}%
    \end{subfigure}
    ~
    \begin{subfigure}[b]{\textwidth}
        \includegraphics[scale=1.3]{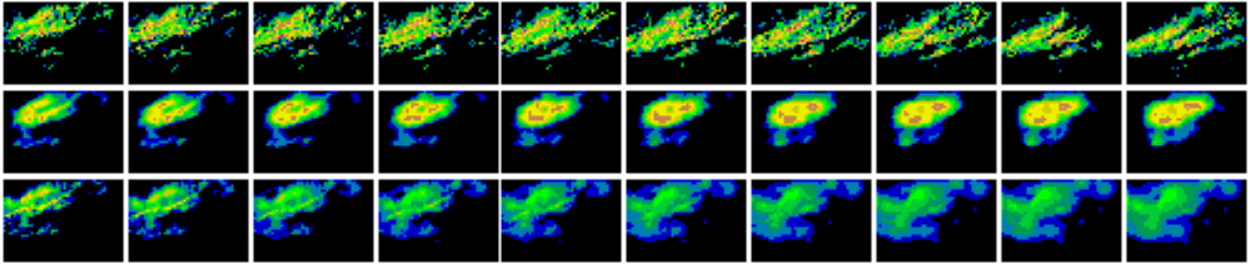}%
    \end{subfigure}    
    \caption{Examples of 5 particular precipitation predictions. \emph{Top row:} Ground truth; \emph{middle row:} stochastic variational frame prediction method; \emph{bottom row:} convolutional LSTM network. All frames are 15 minutes apart, i.e.\ the total lead time is 2 1/2 hours. The results for the stochastic variational frame prediction network represent an average over 10 individual realizations from the model.}\label{fig:IndividualRuns}
\end{figure}

In Figure~\ref{fig:IndividualRuns} we present several individual test cases for the two models, which include both advective and convective precipitation. Results for the stochastic variational frame predictor model are averaged over 10 individual realizations from the model. The stochastic variational frame predictor network typically captures the essential evolution of the precipitation fields, while the standard convolutional LSTM network usually either under- or overestimates the development of the precipitation cells. Additionally, the convolutional LSTM network adds considerably more blur to frames further in the future, which is a well-know phenomenon observed in the video frame prediction literature, see e.g.~\cite{dent18a, math15a}.

\section{Discussion}

We have applied a stochastic variational frame predictor network with a learned prior distribution to the precipitation nowcasting problem. We found the network easy to train in practice, in particular compared to the standard convolutional LSTM networks, which requires relatively long training cycles. The model attempts to encode stochastically meaningful information about the next frames by training a frame-dependent prior distribution. For the precipitation nowcasting problem we interpret this as attempting to learn the physics of the current precipitation cells with the aim to better forecast the following frames. The model is able of producting meaningful longer time precipitation predictions, in particular when compared to the standard convolutional LSTM model, which excessively blurs the predicted precipitation fields. 

Another advantage of the stochastic variational model is that it allows one to predict more than just a single possible future for the evolution of precipitation fields. This is important since the precipitation nowcasting problem is not well-posed, meaning there are many possible evolutions for a given set of input radar frames. Being able to incorporate a level of uncertainty of the future development of active weather regions is of great practical importance in numerical weather prediction, which in recent years has increasingly moved towards ensemble prediction strategies. The fully deterministic convolutional LSTM model is not capable of incorporating stochastic variations and thus can only predict one single deterministic future. Exploring the use of the stochastic frame predictor model with learned prior distribution within the setting of ensemble weather prediction is a potentially promising avenue for future research.

\section*{Acknowledgements}

This research was supported, in part, thanks to the Canada Research Chairs and the NSERC Discovery grant programs and by support provided by ACENET (https://www.ace-net.ca/) and Compute Canada (www.computecanada.ca). The author is grateful to Oliver Stueker for technical help regarding the use of the cluster \texttt{Graham}, and to Markus Kerschbaum, Johannes Sachsperger and Martin Steinheimer at Austro Control for providing the radar data.

\bibliographystyle{rop}
\bibliography{bihlo}

\begin{thebibliography}{10}
\providecommand{\url}[1]{\texttt{#1}}
\providecommand{\urlprefix}{URL }
\expandafter\ifx\csname urlstyle\endcsname\relax
  \providecommand{\doi}[1]{doi:\discretionary{}{}{}#1}\else
  \providecommand{\doi}{doi:\discretionary{}{}{}\begingroup
  \urlstyle{rm}\Url}\fi
\providecommand{\eprint}[2][]{\url{#2}}

\bibitem{baba17a}
Babaeizadeh M., Finn C., Erhan D., Campbell R.H. and Levine S., Stochastic
  variational video prediction, \emph{arXiv preprint arXiv:1710.11252}  (2017).

\bibitem{bowl04a}
Bowler N.E., Pierce C.E. and Seed A., Development of a precipitation nowcasting
  algorithm based upon optical flow techniques, \emph{Journal of Hydrology}
  \textbf{288} (2004), 74--91.

\bibitem{chol15a}
Chollet F. \emph{et~al.}, Keras, \url{https://keras.io}, 2015.

\bibitem{dent18a}
Denton E. and Fergus R., Stochastic video generation with a learned prior,
  \emph{arXiv preprint arXiv:1802.07687}  (2018).

\bibitem{finn16a}
Finn C., Goodfellow I. and Levine S., Unsupervised learning for physical
  interaction through video prediction, in \emph{Advances in neural information
  processing systems}, 2016 pp. 64--72.

\bibitem{higg17a}
Higgins I., Matthey L., Pal A., Burgess C., Glorot X., Botvinick M., Mohamed S.
  and Lerchner A., $\beta$-{VAE}: {L}earning basic visual concepts with a
  constrained variational framework, in \emph{International Conference on
  Learning Representations}, vol.~3, vol.~3, 2017 .

\bibitem{kalt15a}
Kaltenboeck R. and Steinheimer M., Radar-based severe storm climatology for
  {A}ustrian complex orography related to vertical wind shear and atmospheric
  instability, \emph{Atmospheric Research} \textbf{158} (2015), 216--230.

\bibitem{king14a}
Kingma D.P. and Ba J., Adam: {A} method for stochastic optimization,
  \emph{arXiv preprint arXiv:1412.6980}  (2014).

\bibitem{king13a}
Kingma D.P. and Welling M., Auto-encoding variational {B}ayes, \emph{arXiv
  preprint arXiv:1312.6114}  (2013).

\bibitem{lee18a}
Lee A.X., Zhang R., Ebert F., Abbeel P., Finn C. and Levine S., Stochastic
  adversarial video prediction, \emph{arXiv preprint arXiv:1804.01523}  (2018).

\bibitem{lin05a}
Lin C., Vasi{\'c} S., Kilambi A., Turner B. and Zawadzki I., Precipitation
  forecast skill of numerical weather prediction models and radar nowcasts,
  \emph{Geophysical research letters} \textbf{32} (2005).

\bibitem{marsh48a}
Marshall J.S. and Palmer W.M.K., The distribution of raindrops with size,
  \emph{Journal of meteorology} \textbf{5} (1948), 165--166.

\bibitem{math15a}
Mathieu M., Couprie C. and LeCun Y., Deep multi-scale video prediction beyond
  mean square error, \emph{arXiv preprint arXiv:1511.05440}  (2015).

\bibitem{xing15a}
Shi X., Chen Z., Wang H., Yeung D.Y., Wong W.K. and Woo W.c., Convolutional
  {LSTM} network: A machine learning approach for precipitation nowcasting, in
  \emph{Advances in neural information processing systems}, 2015 pp. 802--810.

\bibitem{shi17a}
Shi X., Gao Z., Lausen L., Wang H., Yeung D.Y., Wong W.k. and Woo W.c., Deep
  learning for precipitation nowcasting: {A} benchmark and a new model, in
  \emph{Advances in Neural Information Processing Systems}, 2017 pp.
  5617--5627.

\bibitem{sriv15a}
Srivastava N., Mansimov E. and Salakhudinov R., Unsupervised learning of video
  representations using lstms, in \emph{International conference on machine
  learning}, 2015 pp. 843--852.

\bibitem{vond17a}
Vondrick C. and Torralba A., Generating the future with adversarial
  transformers, in \emph{Proceedings of the IEEE Conference on Computer Vision
  and Pattern Recognition}, 2017 pp. 1020--1028.

\bibitem{wils79a}
Wilson J.W. and Brandes E.A., Radar measurement of rainfall—a summary,
  \emph{Bulletin of the American Meteorological Society} \textbf{60} (1979),
  1048--1060.

\bibitem{xue16a}
Xue T., Wu J., Bouman K. and Freeman B., Visual dynamics: {P}robabilistic
  future frame synthesis via cross convolutional networks, in \emph{Advances in
  Neural Information Processing Systems}, 2016 pp. 91--99.

\bibitem{zahr12a}
Zahraei A., Hsu K.l., Sorooshian S., Gourley J., Lakshmanan V., Hong Y. and
  Bellerby T., Quantitative precipitation nowcasting: a {L}agrangian
  pixel-based approach, \emph{Atmospheric Research} \textbf{118} (2012),
  418--434.

\end{thebibliography}

\end{document}